\pdfoutput=1

\documentclass[11pt]{article}

\usepackage[final]{acl}

\usepackage{times}
\usepackage{latexsym}

\usepackage[T1]{fontenc}

\usepackage[utf8]{inputenc}

\usepackage{microtype}

\usepackage{inconsolata}

\usepackage{graphicx}

\usepackage{multirow}
    \usepackage{graphicx} 
    \usepackage{bm} 
    \usepackage{booktabs}
    \usepackage{colortbl}

\usepackage{xcolor}
\usepackage{makecell}
    
\usepackage{listings}
\lstdefinestyle{prompt}{
    basicstyle=\ttfamily\footnotesize,
    frame = single,
    frameround=tttt,
    backgroundcolor=\color{yellow!10},
    escapechar=\+,
    breakautoindent=false, 
    breaklines=true, 
    breakindent=0pt,
}
\lstdefinestyle{cypher}{
    basicstyle=\ttfamily\small,  
    breaklines=true,          
    frame=none              
}

\title{ZOGRASCOPE: A New Benchmark \\for Semantic Parsing over Property Graphs}

\author{Francesco Cazzaro\footnotemark[2], Justin Kleindienst\footnotemark[4], Sofia Márquez Gomez\footnotemark[4], Ariadna Quattoni\footnotemark[2]\footnotemark[4]\\
  \footnotemark[2]Universitat Politècnica de Catalunya, Barcelona, Spain \\
  \footnotemark[4]dMetrics, Jackson Wyoming USA \\
  \texttt{francesco.cazzaro@upc.edu, aquattoni@cs.upc.edu} \\ \texttt{\{justin.kleindienst, sofia.marquez, ariadna.quattoni\}@dmetrics.com} }

\begin{document}
\maketitle
\begin{abstract}
In recent years, the need for natural language interfaces to knowledge graphs has become increasingly important since they enable easy and efficient access to the information contained in them. In particular, property graphs (PGs) have seen increased adoption as a means of representing complex structured information. Despite their growing popularity in industry, PGs remain relatively underrepresented in semantic parsing research with a lack of resources for evaluation. To address this gap, we introduce \textsc{ZOGRASCOPE}, a benchmark designed specifically for PGs and queries written in Cypher. Our benchmark includes a diverse set of manually annotated queries of varying complexity and is organized into three partitions: iid, compositional and length. We complement this paper with a set of experiments that test the performance of different LLMs in a variety of learning settings.
\end{abstract}

\section{Introduction}

Knowledge graphs (KGs) have become increasingly prominent for representing complex relations between entities across domains and several graph data models have been proposed. Accessing the information contained in a PG typically requires significant expertise in query languages such as Cypher \citep{10.1145/3183713.3190657} or Gremlin \citep{10.1145/2815072.2815073}. To reduce this barrier, natural language interfaces to KGs have emerged as a critical tool. This has driven research in semantic parsing and knowledge-based question answering, which aims to translate natural language questions into executable graph queries \citep{gu-su-2022-arcaneqa,li-etal-2023-shot, agarwal-etal-2024-bring, liang2024aligninglargelanguagemodels, li2024unioqaunifiedframeworkknowledge}. Despite advances, the task remains challenging due to the inherent complexity of natural language and the variability of graph schemas \citep{groschwitz-etal-2023-amr, li-etal-2023-slog}.

Annotated data is essential for advancing development in the area, but acquiring it is both expensive and labor-intensive. Besides, most available benchmarks have focused primarily on RDF-style graphs. Unlike RDF, property graph nodes can have an arbitrary list of associated properties enabling the specification of rich and semantically intuitive data models. PGs clearly distinguish between entity properties and entity relations, whereas RDF graphs conflate the two concepts. Resources for PGs remain severely limited; many available Cypher benchmarks are either synthetically generated without supervision or adapted from non-property graph sources.

To address these challenges, we introduce \textsc{ZOGRASCOPE}, a benchmark designed for property graphs built from an open-access crime investigation graph using Cypher queries. The dataset consists of annotated pairs of natural language questions and Cypher queries. Unlike most prior work, each example is manually annotated by experts. LLMs were used only for intermediate paraphrasing to enhance lexical variety, with all outputs manually verified for correctness. \textsc{ZOGRASCOPE} comprises  $5$k samples, featuring a range of query types, and includes iid, compositional and length train/test partitions. In addition, the paper includes a wide range of  experiments to asses LLMs performance in a variety of learning settings. We hope that \textsc{ZOGRASCOPE} will serve as a valuable resource for advancing research in semantic parsing over property graphs.

We release \textsc{ZOGRASCOPE} at \url{https://github.com/interact-erc/ZOGRASCOPE}.

\section{Related Work}

\begin{table*}[]
\begin{center}
\begin{tabular}{@{}l@{}}

\toprule
\textcolor{gray}{\small Query} \\
\begin{minipage}{0.9\linewidth}
\begin{lstlisting}[style=cypher]
MATCH (x0:Person)-[:KNOWS]-(x1:Person)-[:FAMILY_REL]-(x2:Person WHERE x2.surname = "[x2.surname.VALUE]")
RETURN x0.nhs_no
\end{lstlisting} \end{minipage}\\
\midrule
\textcolor{gray}{\small Proto-NL} \\
\parbox[t]{0.9\linewidth}{\begin{small}
 \textit{List all national health service number of People  who knows People that has a family relation with People with last name that is [x2.surname.VALUE]}
\end{small}}
\\
\midrule
\textcolor{gray}{\small Human annotation} \\
\parbox[t]{0.9\linewidth}{\begin{small}
 \textit{List all national health service numbers of people who knows someone that is family related to a person with surname [x2.surname.VALUE]}
\end{small}}
\\
\midrule
\textcolor{gray}{\small Values Instantiation} \\
\parbox[t]{0.9\linewidth}{\begin{small}
 \textit{List all national health service numbers of people who knows someone that is family related to a person with surname Nichols}
\end{small}}
\\
\midrule
\textcolor{gray}{\small LLM paraphrasing} \\
\parbox[t]{0.9\linewidth}{\begin{small}
 \textit{What are the NHS numbers of individuals who know a family member of someone named Nichols?}
\end{small}}
\\
\midrule
\textcolor{gray}{\small Human verification and correction} \\
\parbox[t]{0.9\linewidth}{\begin{small}
 \textit{What are the NHS numbers of individuals who know a family member of someone with last name Nichols?}
\end{small}}
\\
\bottomrule
\end{tabular}
\end{center}
\caption{ Illustration of the annotation process for one \textsc{ZOGRASCOPE} sample}
\label{tab:zograscope_annotation}
\end{table*}

Few benchmarks exist for evaluating semantic parsers on property graphs, most of which are adapted from RDF and tabular datasets, or synthetically generated by LLMs. Using the first approach \citet{zhao2023cyspider} and \citet{tiwari-etal-2025-auto} adapt SQL benchmarks. \citet{feng2024cypherbench} and \citet{nie-etal-2022-graphq} create benchmarks by converting RDF-graphs to property graphs: the former converts a subset of WikiData \citep{10.1145/2629489} while the latter converts MetaQA \citep{Zhang_Dai_Kozareva_Smola_Song_2018}. Following the second approach \citet{tiwari-etal-2025-auto} and \citet{zhong-etal-2025-synthet2c} use LLMs to build pipelines capable of synthetically producing Cypher benchmarks without human supervision.
\citet{ozsoy-etal-2025-text2cypher} gathers synthetically generated samples from various available online resources to construct a benchmark, but the underlying graph to execute queries is absent.

The main difference of our approach with respect to previous works is that we create a manually annotated Cypher benchmark built directly over a native property graph. In contrast to adapted benchmarks, our exhibits all the characteristics that are unique to the property graph data model. 

\citet{zhao2023rel2graph} employ LLMs to build a benchmark directly on top of a PG adding a final human verification step. The main difference with our benchmark creation process is that in our case all the initial annotations are provided by human annotators. Last, the dataset of \citet{DBLP:conf/cikm/GuoLXT022} resembles our methodology with the difference that it is in Chinese while ours is in English.

\section{Dataset Creation}

The \textsc{ZOGRASCOPE} benchmark is based on the publicly available Pole graph\footnote{https://github.com/neo4j-graph-examples/pole}, which contains a total of $61,521$ nodes and $105,840$ edges. It includes $11$ entity classes, $32$ unique properties, and $17$ relation types. The graph focuses on crime investigations, modeling the relationships between people, objects, locations and events. For instance, a query might ask for the name of an officer investigating a particular crime that occurred at a specific location and involved a particular vehicle. Creating the benchmark involves a five step process: 1) We automatically generate variable anonymized Cypher queries. 2) Each anonymized Cypher query is manually annotated by an expert with a natural language realization. 3) For each query we instantiate the values of the anonymized variables in both the Cypher query and its natural language realization. 4) An LLM is used to suggest paraphrases of the instantiated natural language realizations and 5) A human annotator verifies and corrects the instantiated paraphrases.

Table \ref{tab:zograscope_annotation} illustrates the annotation process. We use four human annotators, all proficient in the Cypher language. At any point, the annotator is free to discard a sample. This quality control mechanism ensures that only meaningful, well-formed queries and their corresponding natural language realizations are included in the final dataset.

To generate Cypher queries, we employ an automatic procedure. We start by constructing a tree pattern, which is essentially a small subgraph, where the root node represents the answer, and the other nodes are connected through edges that represent relations. These tree patterns are designed to be consistent with the underlying graph, ensuring that nodes are connected by valid relations. Nodes in the tree can be unconstrained or constrained by property-value pairs. For example, an unconstrained node might correspond to a \textit{Person} and have no property constrain. Each tree pattern can be deterministically mapped to a corresponding Cypher-query and that is how we generate the anonymized Cypher queries. 

For each query, we also associate a deterministically generated proto-natural language sentence realization, this is only used to aid the human annotator who can decide to ignore it if it prefers to only look at the Cypher query. The annotator is then asked to generate a realistic natural language realization of the query. Annotators are provided with the graph schema, which includes information on entity classes, properties, and relations. The result of this step is an anonymized Cypher query and its corresponding anonymized natural language realization. We then instantiate the anonymized variables with valid values from the graph, making sure that the query yields non-empty results. We generate three instantiations per query. Subsequently, to enrich the lexical diversity of the natural language queries we use an LLM (GPT4o) to produce three paraphrases for each instantiation. In the final step a human annotator reviews each paraphrase verifying and correcting them if needed.

The final dataset generated using this approach has around $5$k samples and the query lengths range from $2$ to $5$ nodes. The queries in the benchmark span a variety of aggregation types, including set operations, set size, attribute value sets, maximum and minimum attribute value sets, and both argmax and argmin attribute values. We split the data into training (\textasciitilde$3$k) and test (\textasciitilde$2$k) sets, organizing the test into an iid and a compositional partition. We define a template as a Cypher query with anonymized properties, classes, relations, and values, retaining only structural information. Test samples that correspond to templates not observed in the training set form the compositional split, while test samples corresponding to templates observed in the training set comprise the iid split. We also provide a length-based split: the training set includes queries with $2$–$3$ nodes, while the test set includes those with $4$–$5$ nodes. In table \ref{tab:partition_stats} we present some statistics and Appendix \ref{sec:appendix_examples} shows some examples. 

\begin{table}[]
\begin{center}
\begin{tabular}{lccccc}
\toprule
\textbf{} & \textbf{Tot.} & \textbf{\%2n.} & \textbf{\%3n.} & \textbf{\%4n.} & \textbf{\%5n.} \\ \midrule
\textsc{train}   & $2905$ & $51.9$ & $28.9$ & $15.2$ & $3.9$\\
\textsc{iid}   & $767$ & $50.1$ & $29.3$ & $16.4$ & $4.2$ \\
\textsc{comp}  & $1350$ & $12.7$ & $47.3$ & $34.6$ & $5.4$ \\
\textsc{len}$_{train}$   & $3769$ & $54.8$ & $45.2$ & $0$ & $0$ \\
\textsc{len}$_{test}$   & $1253$ & $0$ & $0$ & $82.5$ & $17.5$ \\
\bottomrule
\end{tabular}
\end{center}
\caption{Total number of samples in each partition and their distribution among different query lengths.}
\label{tab:partition_stats}
\end{table}

\section{Experiments}

\begin{table*}[]
\begin{center}
\begin{tabular}{llccc}
\toprule
& \textbf{Model} & \textbf{iid} & \textbf{compositional} & \textbf{length} \\ \midrule

\multicolumn{1}{l}{\footnotesize\textit{Fine-tuning}} & Mistral 7B & $97.87$ & $74.97$ & $23.46$ \\
& Qwen3 4B & $98.04$ & $72.42$ & $20.19$ \\
& Llama 3.2-3B     & $97.90$ & $70.99$ & $25.88$ \\

\arrayrulecolor{gray}\midrule

\multicolumn{1}{l}{\footnotesize\textit{Dynamic ICL}} & GPT-4o          & $96.74$ & $66.79$ & $50.19$ \\
\multicolumn{1}{l}{} & Mistral 7B       & $95.18$ & $39.95$ & $13.57$ \\
\multicolumn{1}{l}{} & Qwen3 4B         & $94.79$ & $42.10$ & $18.27$ \\
\multicolumn{1}{l}{} & Llama 3.2-3B     & $97.00$ & $31.73$ & $10.93$ \\

\midrule

\multicolumn{1}{l}{\footnotesize\textit{Fixed ICL}} & GPT-4o        & $66.40$ & $63.71$ & $22.66$ \\
\multicolumn{1}{l}{} & Mistral 7B       & $44.44$ & $25.30$ & $11.33$ \\
\multicolumn{1}{l}{} & Qwen3 4B         & $43.27$ & $28.76$ & $12.69$ \\
\multicolumn{1}{l}{} & Llama 3.2-3B     & $31.03$ & $15.49$ & $5.98$ \\

\midrule

\multicolumn{1}{l}{\footnotesize\textit{Zero-shot}} & GPT-4o       & $41.67$ & $32.91$ & $16.28$ \\
\multicolumn{1}{l}{} & Mistral 7B       & $8.85$ & $4.82$ & $0.56$ \\
\multicolumn{1}{l}{} & Qwen3 4B         & $10.41$ & $7.04$ & $0.88$ \\
\multicolumn{1}{l}{} & Llama 3.2-3B     & $3.38$ & $1.48$ & $0.24$ \\

\arrayrulecolor{black}
\bottomrule
\end{tabular}
\end{center}
\caption{Experimental results on each \textsc{ZOGRASCOPE} partition.}
\label{tab:main_results_cypher}
\end{table*}

We ran a series of experiments to examine semantic parsing performance on \textsc{ZOGRASCOPE}. Specifically, we assess different transformer-based models under four distinct settings:

\textbullet\ \textsc{\textbf{Fine-tuning}}: The model is fine-tuned on the benchmark's training set. Details about the hyperparameters are provided in Appendix \ref{sec:appendix_hyperparam}.
    
\textbullet\ \textsc{\textbf{Dynamic ICL}} In-context learning with $5$ examples dynamically retrieved from the training set based on the test question. We follow \citet{agarwal-etal-2024-bring} in using a sentence transformer as encoder and retrieving based on cosine similarity.

\textbullet\ \textsc{\textbf{Fixed ICL}}  In-context learning with $5$ fixed examples randomly sampled from the training set.

\textbullet\ \textsc{\textbf{Zero-shot}} An LLM is prompted to generate the Cypher query without any example.

Note that for the last three settings we include in the LLM prompt the schema description of the graph with information about classes, properties and relations (see Appendix \ref{sec:appendix_prompt}).

We present the results of these evaluations using a range of LLMs including small models that can run locally: \textit{Llama-3.2-3B} \citep{grattafiori2024llama3herdmodels}, \textit{Qwen3-4B} \citep{yang2025qwen3technicalreport}, and \textit{Mistral-7B-v0.1} \citep{jiang2023mistral7b} as well as a larger proprietary model GPT-4o: (\textit{gpt-4o-2024-08-06}).

For the evaluation metric, we use execution accuracy, where a prediction is deemed correct if it yields identical results to the reference query when executed on the knowledge graph. Since we are interested in the semantic parsing task, we run these experiments assuming perfect entity linking by passing the gold entities to the models. In Appendix \ref{sec:appendix_noentity} we provide some results for experiments without gold entity linking. 

Table \ref{tab:main_results_cypher} presents the results of our experiments. As expected, we observe that the iid partition is the easiest, while performance decreases on the compositional partition. The length partition is the most challenging, with models struggling to generalize to longer queries.
Regarding the different learning settings we observe that performance improves as more supervision becomes available. In the zero-shot setting, with no supervision, performance is low, but just by including a few examples accuracy improves substantially. Dynamically selecting examples further improves the accuracy, with the iid partition almost achieving perfect performance. However, the performance of GPT4o on the compositional partition does not exhibit the same type of improvement. Finally, fine-tuning with supervised data remains the most effective approach achieving the best performance for each of the models.

The performance of the different models performance is usually correlated with their size, with larger models faring better. GPT-4o, being substantially larger, outperforms all other models demonstrating better capabilities. However, we observe that fine-tuning smaller models can still achieve better performance than using GPT4o with in-context learning alone. In the length partition this does not occur, but it is really hard for a fine-tuned model to generalize to longer query structures.

\begin{figure}[]
  \centering
  \includegraphics[width=\linewidth]{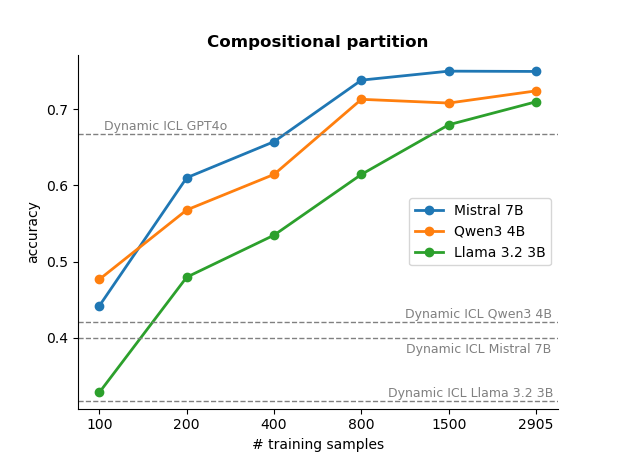}
  \caption{Learning curve for the compositional partition. }
  \label{fig:lc}
\end{figure}

Figure \ref{fig:lc} shows learning curves for the fine-tuned models over the compositional partition. Models are trained on randomly sampled subsets of the training set with sizes of  $100$ $200$ $400$ $800$ and $1500$ examples. We observe a typical learning curve with a sharp rise and then a flattening of the curve.

\section{Conclusions}

In this paper, we introduced \textsc{ZOGRASCOPE}, a novel benchmark designed to evaluate the performance of semantic parsers over Cypher property graphs. Built on the Pole graph, which models crime investigations, the benchmark includes a set of Cypher queries paired with natural language realizations, manually annotated by expert annotators. We hope that \textsc{ZOGRASCOPE} will contribute to advancing research in automatic text-to-Cypher translation, addressing the limited availability of similar resources and fostering progress in the field.

\section*{Limitations}

One limitation of our work concerns the complexity of the query patterns. While we address a range of realistic query structures, certain types such as nested sub-queries are not currently included. In future work, we plan to extend the dataset to cover these more complex cases.
A second limitation regards the entity values in the dataset. Because we instantiate entity values in a second step, even if their wording gets modified, their mapping might still be relatively straightforward. 
Thirdly we acknowledge that the dataset only tackles the english language while it would be useful for these kind of resources to be available for multiple languages.

\section*{Risks and Ethical Considerations}

Our dataset is constructed from a graph that has been rendered publicly available and is based on open data provided by the UK government. Any sensitive information, such as people names, phone numbers, and emails was removed and replaced with fictitious content within the graph prior to its release. As such we believe that our derived dataset does not pose privacy risk nor contain potentially harmful content.

\section*{Acknowledgments}
This project has received funding from the European Research Council (ERC) under the European Union's Horizon 2020 research and innovation programme under grant agreement No 853459. The authors gratefully acknowledge the computer resources at ARTEMISA, funded by the European Union ERDF and Comunitat Valenciana as well as the technical support provided by the Instituto de Física Corpuscular, IFIC (CSIC-UV). This research is supported by a recognition 2021SGR-Cat (01266 LQMC) from AGAUR (Generalitat de Catalunya).

\bibliography{anthology,custom}

\appendix
\clearpage

\section{Hyperparameter details}
\label{sec:appendix_hyperparam}

Regarding the fine-tuning experiments we do an initial parameter selection by reserving a small subset of the data, and then proceed to train on the full dataset for up to $20$ epochs. The parameters that are usually selected are the following: for \textit{Llama-3.2-3B} batch size of $48$ and learning rate of $0.00003$, for \textit{Qwen3-4B} batch size of $48$ and learning rate of $0.00005$ and for \textit{Mistral-7B-v0.1}  batch size of $64$ and learning rate of $0.000005$. We train the model on a single A100 GPU and every result reported is the average of multiple runs with different seeds. In the worst case a training run would usually take $2$ hours.

For the zero-shot and ICLs settings where models are used in inference-only we run experiments on a V100 GPU. For the fixed ICL setting we also repeat the experiments sampling different set of fixed ICL examples and then report the average.

\section{No-Gold Entity Experiments}
\label{sec:appendix_noentity}

\begin{table}[]
\begin{center}
\begin{tabular}{lccc}
\toprule
 & \textbf{iid} & \textbf{compositional} & \textbf{length} \\ \midrule

\multicolumn{1}{l}{\footnotesize\textit{Fine-tuning}}
    & $94.14$ & $68.49$ & $21.38$ \\

\arrayrulecolor{gray}\midrule

\multicolumn{1}{l}{\footnotesize\textit{Dynamic ICL}}
   & $95.05$ & $27.58$ & $10.85$ \\

\midrule

\multicolumn{1}{l}{\footnotesize\textit{Fixed ICL}}
    & $21.93$ & $10.08$ & $5.15$ \\

\midrule

\multicolumn{1}{l}{\footnotesize\textit{Zero-shot}} 
   & $4.16$ & $1.44$ & $0.15$ \\

\arrayrulecolor{black}
\bottomrule
\end{tabular}
\end{center}
\caption{Experimental results for \textit{Llama-3.2-3B} without gold entity linking.}
\label{tab:noentity_results_cypher}
\end{table}

In Table \ref{tab:noentity_results_cypher} we report results for the \textit{Llama-3.2-3B} model without passing the gold entity linking values in input.

\clearpage

\section{ZOGRASCOPE Examples}
\label{sec:appendix_examples}
In Table \ref{tab:zograscope_examples} we display some dataset samples.
\begin{table}[]
\begin{center}
\begin{tabular}{@{}ll@{}}

\toprule
NL &\parbox[t]{1.8\columnwidth}{
\begin{small}
 \textit{What is the number of shoplifting investigations conducted by Police Constables?}
\end{small}
}\\
Query & 
\begin{minipage}{0.8\linewidth}
\begin{lstlisting}[style=cypher]
MATCH (x0:Crime WHERE x0.type = "Shoplifting")-[:INVESTIGATED_BY]-(x1:Officer WHERE x1.rank = "Police Constable")
RETURN COUNT(DISTINCT x0)
\end{lstlisting} \end{minipage} \\
Extra Info &\parbox[t]{1.8\columnwidth}{
\begin{small}
x0.Crime.type:Shoplifting = shoplifting

x1.Officer.rank:Police Constable = Police Constables
\end{small}
}\\
\midrule
NL &\parbox[t]{1.8\columnwidth}{
\begin{small}
 \textit{How long was the last call to the phone of a person linked to a crime on 11/08/2017?}
\end{small}
}\\
Query & 
\begin{minipage}{0.8\linewidth}
\begin{lstlisting}[style=cypher]
MATCH (x0:PhoneCall)-[:CALLER]-(x1:Phone)-[:HAS_PHONE]-(x2:Person)-[:PARTY_TO]-(x3:Crime WHERE x3.date = "11/08/2017")
RETURN x0.call_duration
ORDER BY x0.call_date DESC
LIMIT 1
\end{lstlisting}\end{minipage} \\
Extra Info &\parbox[t]{1.8\columnwidth}{
\begin{small}
x3.Crime.date:11/08/2017 = 11/08/2017
\end{small}
}\\
\midrule
NL &\parbox[t]{1.8\columnwidth}{
\begin{small}
 \textit{Can you list those connected to a crime handled by Officer Rockall, with the investigation finished and no suspect found?}
\end{small}
}\\
Query & 
\begin{minipage}{0.8\linewidth}
\begin{lstlisting}[style=cypher]
MATCH (x0:Person)-[:PARTY_TO]-(x1:Crime WHERE x1.last_outcome = "Investigation complete; no suspect identified")-[:INVESTIGATED_BY]-(x2:Officer WHERE x2.surname = "Rockall")
RETURN x0
\end{lstlisting}\end{minipage} \\
Extra Info &\parbox[t]{1.8\columnwidth}{
\begin{small}
 x1.Crime.last\_outcome:Investigation complete; no suspect identified = investigation finished and no suspect found

x2.Officer.surname:Rockall = Rockall
\end{small}
}\\
\bottomrule
\end{tabular}
\end{center}
\caption{Examples of \textsc{ZOGRASCOPE} data samples.}
\label{tab:zograscope_examples}
\end{table}

\section{LLM Prompt}
\label{sec:appendix_prompt}

\onecolumn

\begin{lstlisting}[style=prompt]
###
Cypher schema:

CLASS: description
Person: People
Location: Locations
Phone: Phone
Email: Email
Officer: Officers
PostCode: PostCode
Area: Areas
PhoneCall: Phone calls
Crime: Crimes
Object: Criminal Objects
Vehicle: Vehicles

PROPERTY: description
year: year
postcode: post code
call_time: time
nhs_no: national health service number
address: address
name: name
phoneNo: phone number
model: model
badge_no: badge number
areaCode: area code
rank: rank
type: type
call_date: call date
call_duration: call duration (seconds)
email_address: email
make: car brand
date: date
surname: last name
code: code
last_outcome: processing status
age: age

RELATION: description - domain -> range
CURRENT_ADDRESS: that lives in - Person -> Location
HAS_PHONE: which has - Person -> Phone
HAS_EMAIL: which has - Person -> Email
KNOWS_SN: that is friends with - Person -> Person
KNOWS: who knows - Person -> Person
HAS_POSTCODE: which has - Location -> PostCode
POSTCODE_IN_AREA: that is in - PostCode -> Area
INVOLVED_IN: that is involved in - Vehicle -> Crime
CALLER: that were made to - PhoneCall -> Phone
CALLED: that were received a - PhoneCall -> Phone
KNOWS_PHONE: knows the phone of - Person -> Person
OCCURRED_AT: that occurred at - Crime -> Location
INVESTIGATED_BY: that is investigated by - Crime -> Officer
PARTY_TO: which is involved in - Person -> Crime
FAMILY_REL: that has a family relation with - Person -> Person
KNOWS_LW: that lives with - Person -> Person
LOCATION_IN_AREA: that is included - Location -> Area
###

# Task
Give me the cypher neo4j query for the following question. (Answer with the query only, do not add anything else).

[[ICL EXAMPLES IF AVAILABLE]]

# Sample
Question: [[QUESTION]]

Query:
\end{lstlisting}
\twocolumn

\end{document}